\documentclass[runningheads]{llncs}

\usepackage{eccv}

\usepackage{eccvabbrv}

\usepackage{graphicx}
\usepackage{booktabs}
\usepackage{algorithm}
\usepackage{algpseudocode}
\usepackage{multirow}
\usepackage{array}
\usepackage{pifont}
\usepackage{subcaption}
\usepackage{enumitem}
\usepackage[accsupp]{axessibility}  % Improves PDF readability for those with disabilities.
\usepackage{marvosym}
\newcommand{\emailmark}{\textsuperscript{(\Letter)}}

\usepackage{hyperref}

\usepackage{orcidlink}

\usepackage{marvosym}

\newcommand\blfootnotetext[1]{%
  \begingroup
  \renewcommand\thefootnote{}%
  \footnotetext{#1}%
  \addtocounter{footnote}{-1}%
  \endgroup
}

\begin{document}\sloppy

% \title{Next Geometric Consistency for Autoregressive 3D Point Cloud Generation} 
\title{Learning to Tessellate: Point Cloud Generation via Recursive Spectral Partitioning}

\titlerunning{Point Cloud Generation via Recursive Spectral Partitioning}

\author{
Monan Sun\inst{1,2,4}\orcidlink{}  \and
Bangzhen Liu\inst{3}\emailmark\orcidlink{0000-0001-6621-0594}  \and
Huaidong Zhang\inst{1}\emailmark\orcidlink{}  \and
Shengfeng He \inst{4}\orcidlink{0000-0002-3802-4644}
}

\authorrunning{M.~Sun et al.}

\institute{
South China University of Technology, China\\ \and
University of Chinese Academy of Sciences, China \\\and
City University of Hong Kong, Hong Kong SAR, China \\ \and
Singapore Management University, Singapore \\
\email{mnsun94@gmail.com}, 
\email{bangzliu@cityu.edu.hk} \\
\email{huaidongz@scut.edu.cn}, 
\email{shengfenghe@smu.edu.sg} \\
\url{https://huggingface.co/Mo-nan/PointRSP}
}

\maketitle
\blfootnotetext{\Letter\ Corresponding authors.}

\begin{figure}[h]
  \centering
  \includegraphics[width=\linewidth]{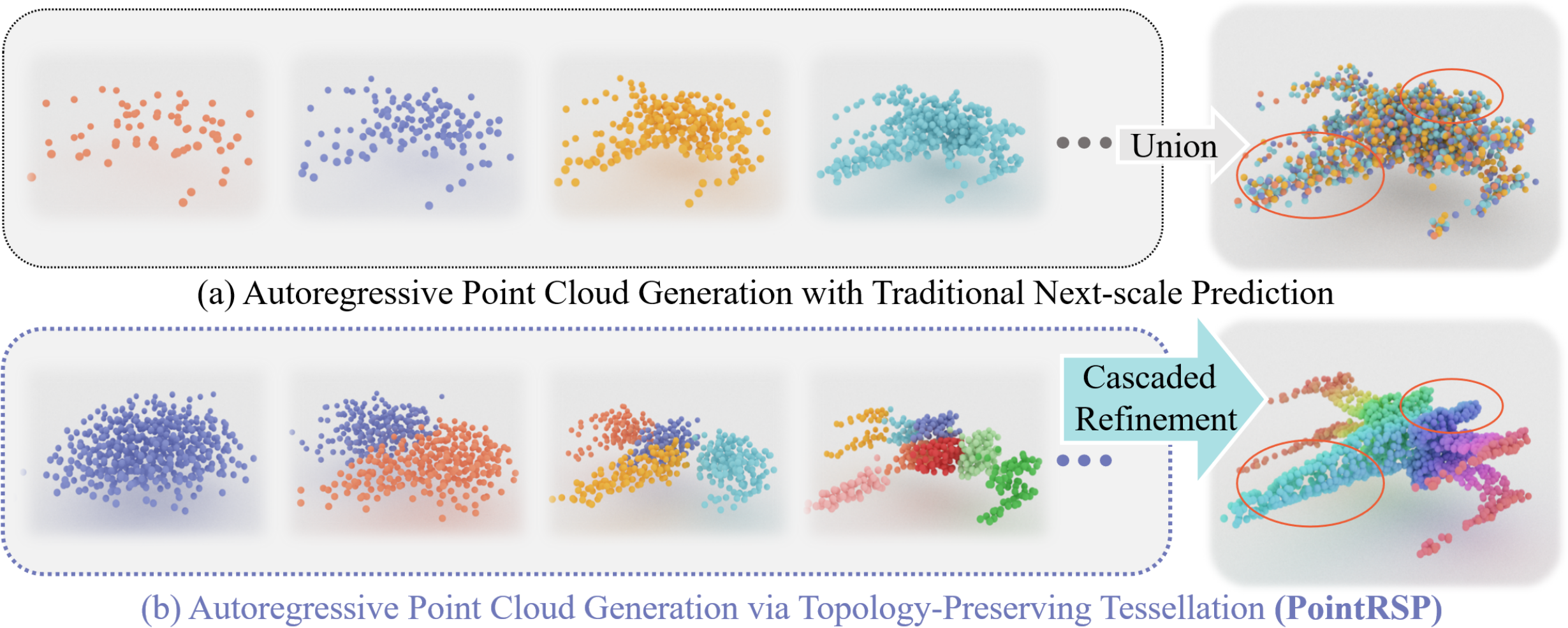}
  \caption{Comparison between PointRSP and conventional fragmented autoregressive models. Traditional methods~(a) generate shapes by merging independently generated subsets, whereas ours~(b) follows a topology-preserving tessellation process.}
  \label{fig:0}
\end{figure}

\begin{abstract}
Autoregressive models have emerged as an effective paradigm for point cloud generation. However, most existing approaches rely on heuristic tokenization strategies, such as spatial sorting or stochastic downsampling, which often disrupt intrinsic point cloud topology and weaken the structural coherence of the generated shapes. 
In this paper, we present PointRSP, an autoregressive framework that reformulates point cloud generation as a topology-preserving tessellation process via recursive spectral partitioning. Instead of constructing token sequences heuristically, we introduce a topology-aware partitioning autoencoder that decomposes an unstructured point cloud into a non-balanced binary tree through a hybrid recursive spectral partitioning strategy. This hierarchical representation provides a deterministic geometric blueprint that preserves topological relationships while capturing multiscale structural dependencies within a quantized latent space.
To synthesize shapes in this space, we propose a dual-stream cascaded generator that jointly models structural evolution and feature synthesis. In addition, we design a geometry-calibrated positional encoding mechanism that anchors latent embeddings using multi-scale structural centers, which stabilizes cascaded generation during the early stages of structural formation. Extensive experiments show that PointRSP achieves state-of-the-art performance in generation quality and diversity, demonstrating strong generalization across complex 3D topologies. 
\end{abstract}

\section{Introduction}
\label{sec:intro}

Synthesizing high-quality point clouds is pivotal for modern 3D intelligent systems, facilitating the modeling of complex 3D distributions and driving advancements in embodied perception~\cite{wang2023robogen,wang2024embodiedscan} and reconstruction~\cite{cheng2022novel,wu2025embodiedocc}. While diffusion-based models have emerged as the prevailing paradigm~\cite{luo2021diffusion,zhou20213d,wang2025pdt,vahdat2022lion,ren2024tiger,zheng2025recdreamer,Tao_2025_ICCV}, their reliance on holistic denoising within a global geometric context often precludes the preservation of fine-grained local features, resulting in artifacts that violate the underlying topological structures. 

Autoregressive (AR) models~\cite{sun2020pointgrow,lee2022autoregressive,cheng2022autoregressive,wu2024point, sun2024autoregressive} offer an alternative by explicitly modeling long-range geometric dependencies through sequential prediction. Despite their efficiency, effectively tokenizing point clouds while respecting their intrinsic geometric topology remains a great challenge. Unlike regular grids, point clouds are unstructured and lack a canonical ordering aligned with the underlying shape manifold. Existing schemes typically rely on heuristic discretization, such as spatial coordinate sorting~\cite{yan2022shapeformer} or space-filling curves~\cite{chen2023pointgpt}, which often fail to capture manifold continuity, resulting in incoherent local details and fragmented structural outputs. Therefore, we posit that it is more natural to view point cloud tokenization as a structural decomposition problem, where the model learns to implicitly tessellate 3D space in a manner consistent with topology.

Recently, next-scale prediction AR models have advanced this domain~\cite{zhang2024g3pt, meng2025pointnsp, molodyk2025mfm} by tokenizing point clouds along an axis of increasing geometric detail (Figure~\ref{fig:0}(a)). 
While this paradigm alleviates the restrictive unidirectional dependencies of fixed linear orderings, the scale definitions remain largely heuristic and stochastic. Consequently, these frameworks struggle to preserve stable spatial or structural correspondence across varying levels of detail. 
To establish robust structural priors, a topology-aware strategy is required to partition the geometric structure rather than relying on heuristic downsampling.

In this context, local geometric relations are critical for learning expressive 3D representations. This principle has been validated by graph-based point cloud understanding architectures such as DGCNN~\cite{wang2019dynamic}. Since these models successfully learn from local topology via dynamically constructed neighborhood graphs, a compelling question naturally arises: can we develop an AR generative model that tokenizes 3D space by explicitly preserving graph connectivity? 
By leveraging graph clustering, we can explicitly decompose a global point cloud into a series of locally connected groups, preserving the structural integrity throughout the generation process.

Motivated by the above insights, we introduce PointRSP, an autoregressive generation framework that reformulates point cloud generation as a topology-preserving tessellation process driven by recursive spectral partitioning. 
PointRSP consists of two components: a Topology-Aware Partitioning Autoencoder and a Dual-Stream Autoregressive Generator. To construct a robust geometric blueprint, the autoencoder represents point clouds as nearest neighbor topology graphs and applies a hybrid recursive spectral partitioning strategy. This strategy combines a greedy merging stage to capture local geometric details and a top-down spectral bisection stage to preserve global manifold topology. By tracking these sequential partition decisions, we linearize the unstructured point set into a hierarchical, non-balanced binary tree. This explicit tree structure acts as a deterministic prior, directly guiding the structural quantization of local point features to construct a multiscale, topology-aware latent space.

To synthesize point clouds within this latent space, our Dual-Stream Generator employs a cascaded architecture where structural priors explicitly dictate content synthesis. At each generation step, a structural predictor models the binary branching patterns of the tessellation tree, while a feature predictor generates residual coordinates conditioned on this evolving structure. To overcome the cold-start instability inherent in early cascaded generation, where reconstructed coordinate cues are highly unreliable, we propose a Geometry-Calibrated Positional Encoding mechanism. This module anchors early latent embeddings by calibrating them against multiscale geometric centers derived directly from the generated structural hierarchy. By tightly intertwining structural prediction with feature synthesis, PointRSP achieves stable, interpretable, and topologically faithful 3D generation.

Our contributions are summarized as fourfold:
\begin{enumerate}[leftmargin=*, itemsep=0pt, topsep=0pt]
\item We reformulate point cloud tokenization as a principled structural decomposition. By framing generation as a manifold-consistent tessellation, the generative sequence is governed by intrinsic geometric topology, mitigating the structural fragmentation caused by heuristic ordering strategies.
\item We propose a topology-aware partitioning autoencoder that constructs a hierarchical binary tree using hybrid recursive spectral partitioning, capturing multiscale geometric dependencies within a quantized latent space.
\item We introduce a dual-stream cascaded generator that jointly models structural evolution and feature synthesis, together with a geometry-calibrated positional encoding that stabilizes early-stage generation.
\item Extensive experiments show that PointRSP achieves state-of-the-art performance in generation quality and diversity, while faithful to 3D topologies of complex 3D shapes.
\end{enumerate}

\section{Related Work}

\textbf{Point Cloud Generation.}
Existing point cloud generation methods can be broadly categorized into continuous distribution models and discrete autoregressive frameworks. Early continuous approaches, such as PointFlow~\cite{yang2019pointflow}, employ normalizing flows to model latent geometric distributions. More recently, diffusion models have become the dominant paradigm~\cite{luo2021diffusion,zhou20213d,vahdat2022lion,ren2024tiger,wang2025pdt,liu2025genpoly}, generating shapes by iteratively reversing a noise corruption process in continuous or latent spaces. While these methods achieve high fidelity, the sparsity of point clouds combined with high-dimensional sampling trajectories results in substantial computational cost. Moreover, their reliance on global denoising contexts may limit the preservation of fine-grained geometric topology.

Autoregressive models provide a more efficient alternative by flattening unordered point sets into discrete sequences for step-wise prediction. A key challenge is defining a meaningful ordering for inherently permutation-invariant data. Prior work addresses this through heuristic linearization strategies, including voxel-based spatial traversal~\cite{yan2022shapeformer}, canonical patch serialization~\cite{cheng2022autoregressive}, and space-filling curves~\cite{chen2023pointgpt}. Although these approaches reduce certain directional dependencies, the imposed orderings often disrupt intrinsic manifold continuity, leading to structural fragmentation.
Recent next-scale prediction methods, such as PointNSP~\cite{meng2025pointnsp}, extend visual autoregressive modeling~\cite{tian2024visual} by progressively predicting subsets of points at increasing resolutions. This strategy improves stability compared with strict linear prediction. However, stochastic or heuristic subset sampling can weaken geometric correspondence across scales. In contrast, PointRSP reformulates the autoregressive sequence through a deterministic structural hierarchy derived from recursive spectral partitioning, enabling topology-consistent generation across the entire process.

\noindent\textbf{Visual Autoregressive Generation Models.}
Visual autoregressive models synthesize complex outputs by iteratively generating discrete components. These approaches are typically divided into sequential and cascaded paradigms. Sequential models~\cite{chen2020generative,esser2021taming} represent images as collections of tokens or patches generated in a fixed order. Cascaded models~\cite{ho2022cascaded,razavi2019generating} instead adopt a hierarchical coarse-to-fine strategy, first producing low-resolution representations and progressively refining them to higher resolutions. Recent work further explores factorizing generation targets through token granularity~\cite{bachmann2025flextok,gao2025d} or frequency-domain representations~\cite{yu2025frequency} to improve efficiency.

Despite the efficiency advantages of cascaded generation, most autoregressive methods for 3D point clouds~\cite{yan2022shapeformer,cheng2022autoregressive,chen2023pointgpt}, including subset-based approaches such as PointNSP~\cite{meng2025pointnsp}, largely follow the sequential paradigm. In these frameworks, token groups across stages remain discrete and independent, imposing strong constraints on sampling strategies. In contrast, PointRSP introduces a cascaded autoregressive framework guided by recursive spectral partitioning. The resulting topology-aware hierarchical sequence preserves structural dependencies across stages, improving both generation efficiency and geometric fidelity.

\begin{figure}[tb]
  \centering
  \includegraphics[width=1\linewidth]{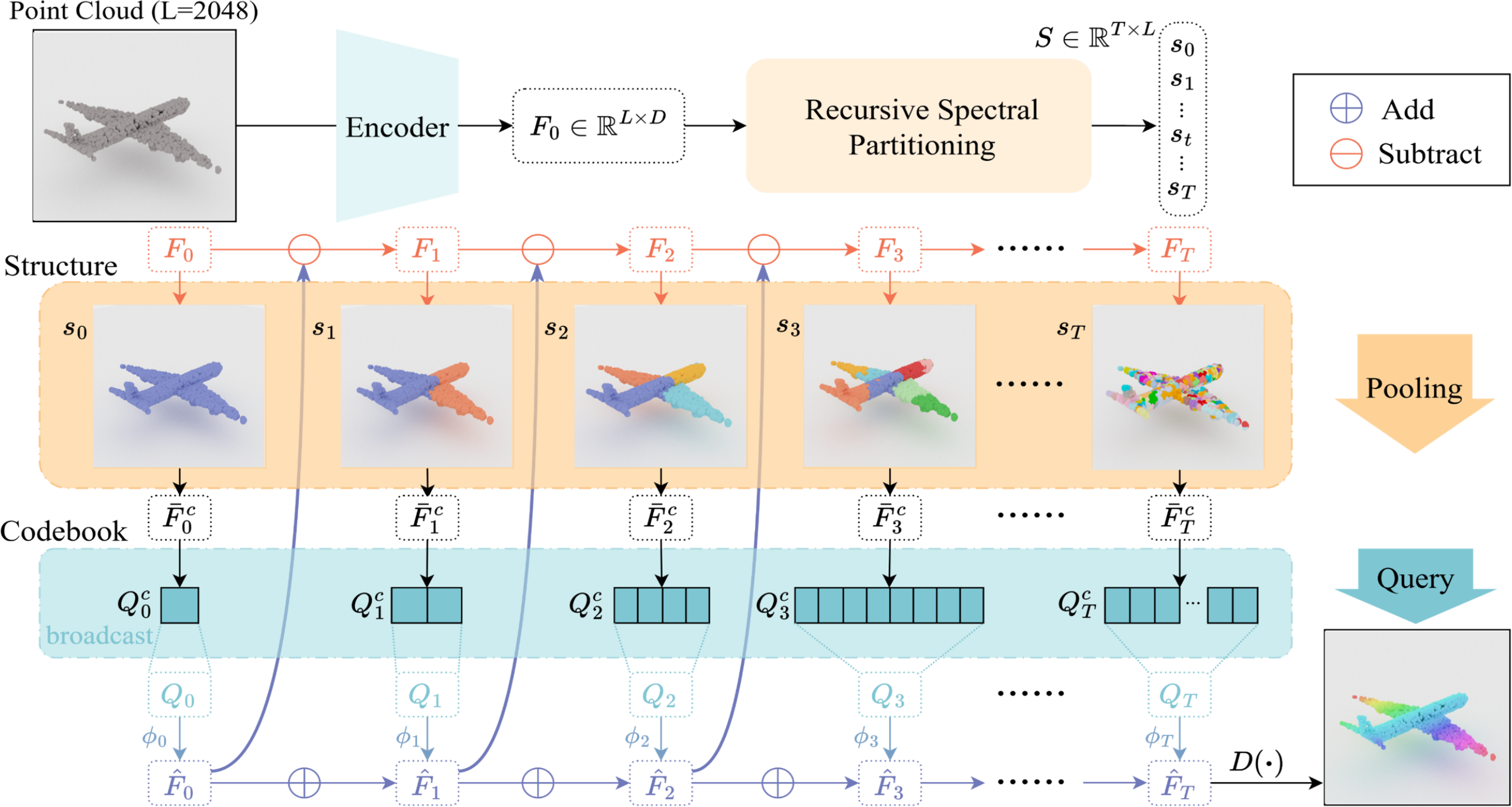}
  \caption{Overview of the Topology-Aware Partitioning Autoencoder. The framework extracts a hierarchical feature sequence, applying a topology-preserving tessellated quantization via a recursive spectral partitioning strategy. By dynamically quantizing features within disjoint geometric subsets dictated by a deterministic structural blueprint, the model effectively preserves intrinsic topology during reconstruction.
  }
  \label{fig:1}
\end{figure}

\section{Method} 

Standard autoregressive frameworks~\cite{van2017neural,sun2020pointgrow} synthesize 3D shapes by decoding discrete tokens according to a predefined, heuristic ordering. However, such artificial linearizations fundamentally disrupt the intrinsic manifold continuity of unstructured point clouds. To address this, we propose PointRSP, which reformulates point cloud tokenization and generation as a topology-preserving tessellation process driven by recursive spectral partitioning.
Accordingly, Section~\ref{sec:vq} details the Topology-Aware Partitioning Autoencoder, which decomposes the point cloud into a non-balanced binary tree to serve as a deterministic geometric blueprint for latent space quantization. Section~\ref{sec:gen} introduces the Dual-Stream Autoregressive Generator, which employs a geometry-calibrated cascaded architecture to jointly model structural evolution and continuous feature synthesis.

\subsection{Topology-Aware Partitioning Autoencoder\label{sec:vq}}

Starting from an input point cloud $X=\{ x_i\}_{i=1}^{L}$ with $L$ points, where $x_i\in\mathbb{R}^3$ is the coordinate in the Euclidean space. We first employ a sparse convolutional encoder~\cite{yan2018second} to obtain latent point feature $F_0\in\mathbb{R}^{L\times D}$, where $D$ denotes the latent dimension. 
The tessellation of the point features follows a cascaded design, where we establish $T$ scale levels to produce a coarse-to-fine global shape feature hierarchy $\mathbf{F}=\{ F_t\}_{t=0}^{T}$. Figure~\ref{fig:1} illustrates the overall autoencoding and quantization pipeline. Through our designed \textit{topology-preserving tessellated quantization}, at each level $t$, every token $f(i,t) \in F_t$ is explicitly associated with a structural label $s{(i,t)}$. These labels collectively form the topology annotation $S$ via a hybird recursive spectral partitioning strategy, which acts as the deterministic geometric blueprint for the subsequent training.

At scale $t$, the structural label $S_t$ assigns each token to a disjoint geometry subset identified by a unique label $c \in \{1,\ldots, C_t\}$. The $C_t$ here denotes the possible number of subset at scale $t$. We then pool the token features within each subset to obtain structure-level representations:

\begin{equation}
  F_t^c = \{f(i,t) \mid s(i,t) = c\}, \quad \bar{F}_t^c = \frac{1}{|F_t^c|} \sum_{f \in F_t^c} f,
  \label{eq:1_2}
\end{equation}
where $|F_t^c|$ denotes the cardinality of the subset with label $c$.
Subsequently, the pooled feature $\bar{F}_{t}^{c}$ serves as the atomic quantization target. This representation is mapped to a discrete code via a nearest-neighbor search within a learnable codebook $\mathbb{Z}$. The selected discrete code is subsequently broadcast back to all tokens sharing the label $c$, yielding a quantized sequence $Q_t \in \mathbb{R}^{L \times D}$. When $t > 0$, we proceed the residual features at scale $t\!-\!1$ to its next scale for stable learning of the quantized features~\cite{tian2024visual}. 
To encourage the progressive refinement of residual geometric details at finer scales, we process the quantized embeddings $Q_t$ through a lightweight mapping network $\phi_t(\cdot)$ to generate the reconstructed features $\hat{F}_t$. The final 3D shape is reconstructed by aggregating these quantized features across all $T$ levels and decoding the accumulated representation via a MLP decoder $D$:
\begin{equation}
  \hat{F}_t = \phi_t(Q_t), \quad \hat{X} = D\left(\sum_{t=0}^T \hat{F}_t\right).
  \label{eq:4}
\end{equation}

\begin{figure}[tb]
  \centering
  \includegraphics[width=\linewidth]{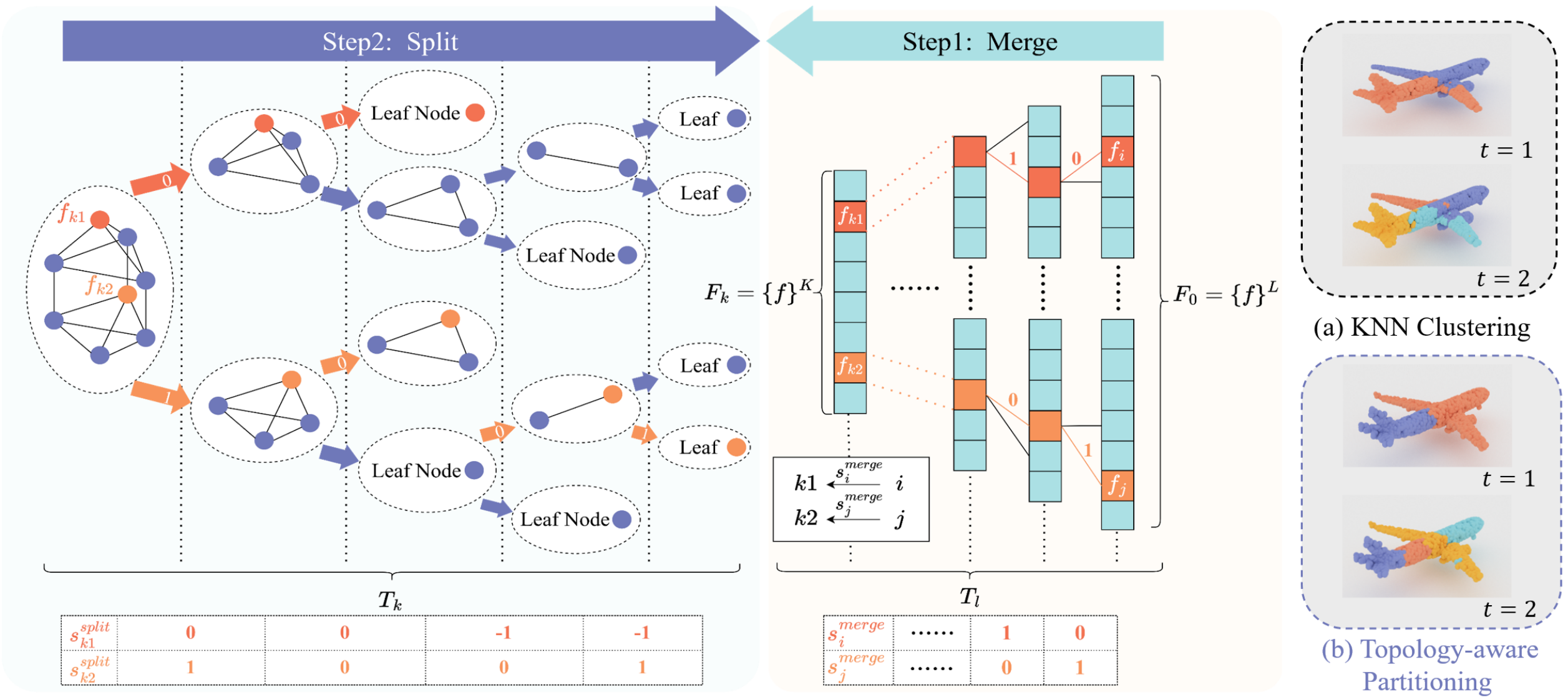}
  \caption{
  Visualization of the recursive spectral partitioning strategy. The geometric blueprint is constructed through a hybrid approach: a bottom-up greedy merge stage ($T_l$) captures fine-grained local details, while a top-down spectral split stage ($T_k$) preserves global topology. Compared to (a) standard KNN-style balanced clustering, (b) our approach explicitly accommodates the non-balanced nature of 3D geometries to maintain strict boundary integrity.
  }
  \label{fig:2}
\end{figure}

\noindent\textbf{Topology-Preserving Tessellated Quantization.\label{sec:gcsm}}
Existing hierarchical quantizers, such as NVG~\cite{wang2025next}, initialize a binary sequence matrix $S^{(2)}\in\{0,1\}^{L\times T}$ to record token partition decisions. Here the partition decision of token $i$ at scale $t$ is denoted as $s^{(2)}(i,t)$. Structural labels are obtained by converting these binary codes into decimal indices via operation $\text{Dec}(\cdot)$:
\begin{equation}
    \text{Dec}(s^{(2)}(i,j)) = \sum_{j=0}^{t} s^{(2)}(i,j)\,2^{t-j}. \label{eq:dec}
\end{equation}
This formulation preserves cross-scale hierarchical relations but implicitly enforces balanced spatial bisections through KNN-style clustering, restricting the number of structural labels at stage $t$ to exactly $2^t$. While acceptable for dense 2D grids, uniform splitting is highly restrictive for unstructured point clouds, where geometry is naturally unbalanced across different branches and scales. Forced uniform bisection frequently blurs topological boundaries and erroneously merges structurally unrelated regions (see Figure~\ref{fig:2}(a)).

To accurately respect intrinsic geometric relations, we propose a recursive spectral partitioning strategy based on graph-Laplacian spectral clustering~\cite{belkin2003laplacian,von2007tutorial}. Because token-level spectral partitioning across the entire point cloud is computationally expensive and can yield misaligned structural sequences, we introduce a hierarchical hybrid strategy. As illustrated in Figure~\ref{fig:2}, we decompose the quantization blueprint into a bottom-up greedy merge stage with $T_l$ steps and a top-down spectral split stage with $T_k$ steps, satisfying $T_l+T_k=T$.

In the initial merge stage, we start from the raw feature sequence $F_0$ and iteratively merge the most similar token pairs based on their local feature distances. This coarse-scale greedy merging captures fine-grained local details and reduces the sequence length from $L$ to $K$, bounded by the condition $\log_2 L-\log_2 K=T_l$. We record these bottom-up binary decisions at each step as $S^{\text{merge}} \in \{0,1\}^{L \times T_l}$ and extract the pooled features $F_k$ to serve as computationally efficient nodes for the subsequent global stage.

During the spectral split stage, we construct a nearest-neighbor graph Laplacian over the $K$ merged tokens and perform recursive spectral bipartitioning for $t=0,\ldots,T_k-1$. At each step, we compute the Fiedler vector of the graph Laplacian to capture the dominant geometric variation. Let $v(k,t)$ denote the Fiedler vector value for token $k$ at step $t$. Structural split labels are deterministically assigned based on the sign of $v(k,t)$. Because this topology-driven partitioning yields a non-balanced binary tree, tokens reach terminal leaf states at variable depths. Let $l_k$ denote the effective split sequence length of the $k$-th merged token. We define the global split horizon as $T_k=\max\{l_k\}$ for $k=1,\ldots,K$. The split labels $S^{\text{split}}$ are assigned as:
\begin{equation}
  s^{\text{split}}(k,t) =
  \begin{cases}
  0, & v(k,t) \le 0, \quad t < l_k, \\
  1, & v(k,t) > 0, \quad t < l_k, \\
  -1, & \text{otherwise}.
  \end{cases}
  \label{eq:6}
\end{equation}

Finally, we establish the global topology-preserving structural blueprint $S^{(2)}$ by integrating the binary labels from both the merging and splitting stages. An alignment function $\text{map}(\cdot)$ traces each original point $i$ through the merge hierarchy to its corresponding aggregated token $k$ in the split stage. This alignment guarantees structural consistency across the entire generation process:
\begin{equation}
  s^{(2)}(i,t) =
  \begin{cases}
  s^{\text{split}}(k,t), & t < l_k, \ k = \text{map}(i, S^{\text{merge}}), \\
  s^{\text{merge}}(i, t - l_k), & l_k \le t < l_k + T_l, \\
  0, & \text{otherwise}.
  \end{cases}
  \label{eq:7}
\end{equation}

\noindent\textbf{Training Loss.}
The Topology-Aware Partitioning Autoencoder is optimized end-to-end using a weighted combination of three objectives: a Smooth-$\ell_1$ loss $L^{\text{smooth}}_{\text{L1}}$~\cite{girshick2015fast} between the original points $X$ and the reconstructed $\hat{X}'$, a Chamfer loss $L_{\text{cmf}}$ to enforce global shape-level geometric fidelity, and an improved quantization loss $L_{\text{vq}}$ with better codebook usage~\cite{luo2024open} to ensure latent feature consistency between the initial features $F_0$ and the quantized outputs $\hat{F}$:
\begin{equation}
  \mathcal{L}_{\text{total}} =
  \lambda_{L1} L^{\text{smooth}}_{\text{L1}}(X, \hat{X}) +
   \lambda_{cmf} L_{\text{cmf}}(X, \hat{X}) + 
   \lambda_{vq} L_{\text{vq}}(F_0, \hat{F}_T).
  \label{eq:13}
\end{equation}

\subsection{Dual-Stream Autoregressive Generator\label{sec:gen}}

\begin{figure}[tb]
  \centering
  \includegraphics[width=\linewidth]{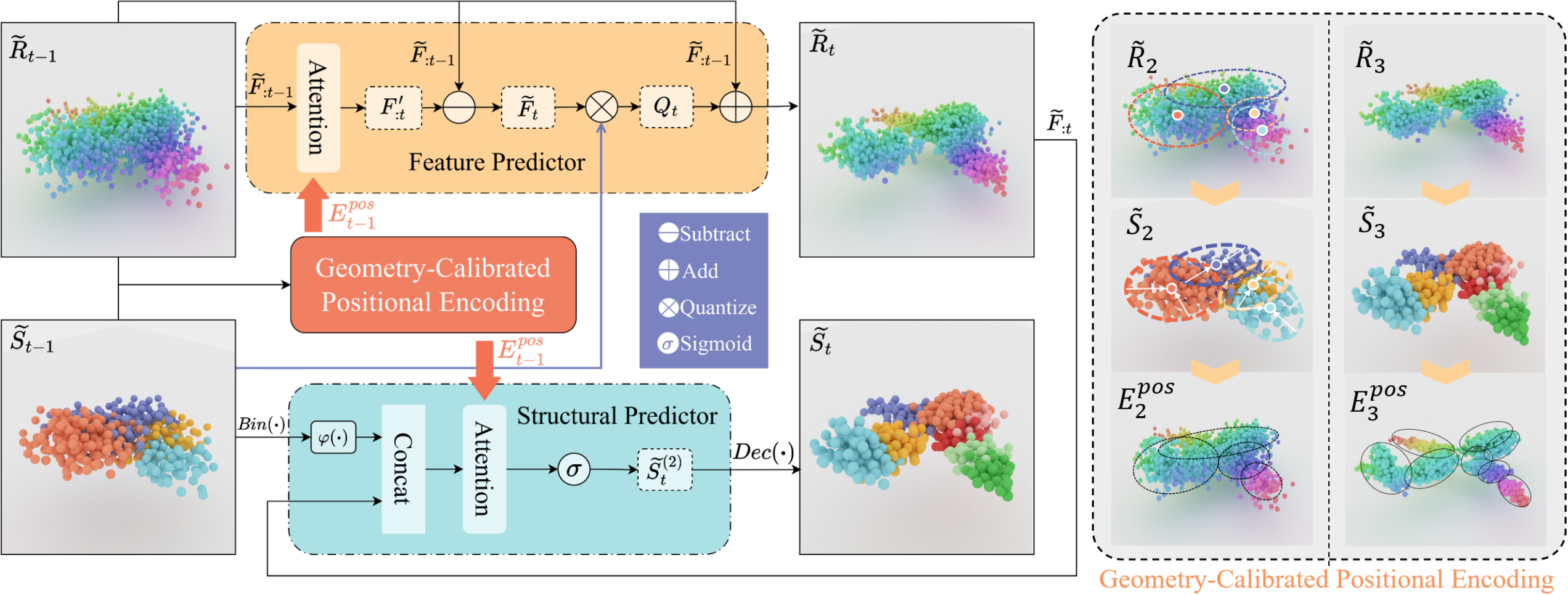}
  \caption{
  Overview of the cascaded generation from scale $t-1$ to $t$. The dual-stream generator employs a structural predictor and a feature predictor, stabilized by a geometry-calibrated positional encoding. The right panel illustrates the dynamic construction of topology-aware positional embeddings to resolve early-stage cold-start instability.
  }
  \label{fig:3}
\end{figure}

Standard autoregressive 3D generators typically encode structural dependencies implicitly via manually crafted attention masks. In contrast, PointRSP adopts a topology-aware dual-stream cascaded paradigm (see Figure~\ref{fig:3}) comprising two explicit modules: a structural predictor that models inter- and intra-scale geometry, and a feature predictor that performs residual feature synthesis conditioned on the predicted structure. By promoting structural prior from an implicit constraint to an interpretable, recursively updated representation, the generation process is directly guided by the evolving geometric blueprint.

Let $\widetilde{F}_t$ denote the generated feature representation at scale $t$, and let $\widetilde{F}_{:t}$ represent the accumulation of generated features from resolution levels $1$ to $t$. Prior to generation, the initial feature sequence at $t=1$ is initialized as a zero vector, and the corresponding structural sequence is initialized by sampling from a standard Gaussian distribution.
During generation, the scale index $t$ iterates from $1$ to $T$. Each cascaded step executes three core operations: computing \textit{geometry-calibrated positional embeddings}, predicting next-level features conditioned on the current structure, and updating the structural representation. Both predictors are built upon self-attention mechanisms and utilize Rotary Position Embedding (RoPE)~\cite{shi2024taming}.
First, the feature representation $\widetilde{F}_{t}$ is predicted from the accumulated history $\widetilde{F}_{:t-1}$ using the feature attention module $\text{Attn}^{f}$:
\begin{equation}
  \widetilde{F}_{t} = \text{Attn}^{f}(\widetilde{F}_{:t-1}) - \widetilde{F}_{:t-1}.
  \label{eq:8}
\end{equation}
These features are then quantized under the guidance of the preceding structural sequence $\widetilde{S}_{t-1}$. We employ a MLP-based projection layer to associate the pooled features with codebook entries, ensuring stable gradient propagation. The accumulated feature sequence for the next step is computed as:
\begin{equation}
  \widetilde{F}_{:t} = \text{Quantize}(\widetilde{F}_{t}, \widetilde{S}_{t-1}) + \widetilde{F}_{:t-1}.
  \label{eq:9}
\end{equation}

To predict the subsequent structural sequence, we concatenate the accumulated features $\widetilde{F}_{:t}$ with the prior structural labels $\widetilde{S}_{t-1}$, enabling the generated feature content to iteratively refine the structural blueprint. A lightweight projection $\varphi$ ensures dimensional compatibility prior to processing by the structural attention module $\text{Attn}^s$:
\begin{equation}
  \widetilde{S}^{(2)}_{t} = \sigma \left( \text{Attn}^s \left( \left[ \widetilde{F}_{:t}, \varphi(\text{Bin}(\widetilde{S}_{t-1})) \right] \right) \right) > 0,
  \label{eq:10}
\end{equation}
where $\text{Bin}(\cdot)$ denotes the inverse process of Eq.~\ref{eq:dec}, $\sigma(\cdot)$ is the sigmoid activation function to squash the output into probabilistic binary decisions that define the structural path of each node. After $T$ recursive steps, the final accumulated feature sequence $\widetilde{F}_{:T}$ is decoded into the generated 3D shape.

\noindent\textbf{Geometry-Calibrated Positional Encoding.\label{sec:sreb}}
Unlike 2D images equipped with regular grid indices, unstructured point clouds lack dense positional priors. Furthermore, collapsing 3D coordinates into a single 1D index space~\cite{ren2024tiger,meng2025pointnsp} severely distorts spatial proximity and undermines local geometric consistency.

To address this, we formulate the positional embedding at scale $t$ as the concatenation of a structural term $E^{s}_{t}$ and a feature-based term $E^{\text{pos}}_{t}$. The structural term is directly derived from the discrete identifiers in $\widetilde{S}_{t}$, reducing embedding collisions by providing unique topological cues. For the feature-based term, we utilize the intermediate reconstructed coordinates $\widetilde{X}_{t-1} \in \mathbb{R}^{L \times 3}$ as geometric anchors, which are decoded from the accumulated features via $\widetilde{X}_{t-1}=D(\widetilde{F}_{:t-1})$.

However, cascaded generation suffers from a ``cold-start" instability. In early stages, the reconstructed coordinates are highly unreliable, making direct coordinate-based positional encoding prone to severe noise. To resolve this, we propose a geometry-calibrated positional encoding that anchors early features using structure-induced geometric centers. Given the predicted structure labels $\widetilde{S}_t$, we compute group centers $P_t \in \mathbb{R}^{M_t \times 3}$, where $M_t$ is the number of distinct structural groups. For a token $i$ with reconstructed coordinates $x(i,t) \in \widetilde{X}_t$, its geometry-calibrated positional embedding $e(i,t) \in E^{\text{pos}}_t$ is defined as:
\begin{equation}
  e(i,t) =
  \begin{cases}
  \alpha \cdot x(i,t) + (1-\alpha) \sum_{j=0}^{t-1} w_j \cdot P \left(\text{map}(i,\widetilde{S}_t), j \right), & t < T_k, \\
  x(i,t), & t \ge T_k.
  \end{cases}
  \label{eq:12}
\end{equation}
Here, $\text{map}(i,\widetilde{S}_t)$ returns the structural group index of token $i$. The hyperparameter $\alpha$ controls the reliance on the raw reconstructed coordinates. To prioritize finer geometric details, the weight $w_j$ for the historical geometric center at scale $j$ increases exponentially ($w_{j+1} = 2w_j$), constrained by $\sum_{j=0}^{t-1} w_j = 1$. When $t < T_k$, the representation is robustly anchored by these multi-scale structural centers. Once generation reaches the split horizon ($t \ge T_k$), the encoding transitions exclusively to $x(i,t)$ since it is deemed geometrically stable.

\noindent\textbf{Training Loss.}
During the generation phase, both the feature predictor and the structural predictor are supervised end-to-end using a Mean Squared Error (MSE) objective. We measure the deviation between the generated features $\widetilde{F}_{:t}$ and structures $\widetilde{S}^{(2)}_t$ against their corresponding quantized ground-truth representations extracted by the autoencoder:
\begin{equation}
  \mathcal{L}_{\text{Gen}} =
  \frac{1}{T}\sum_{t=1}^{T} L_{\text{mse}}\!\left(\hat{F}_t,\widetilde{F}_{:t}\right)
  + \frac{1}{T_k}\sum_{t=1}^{T_k} L_{\text{mse}}\!\left(S^{(2)}_t,\widetilde{S}^{(2)}_t\right).
  \label{eq:14}
\end{equation}
Crucially, we restrict structural supervision exclusively to the spectral split stage ($t=1,\ldots,T_k$). This targeted supervision prevents the model from overfitting to local variations and stabilizes the training of the structural predictor during the critical early stages of cascaded generation.

\begin{figure}[t]
  \centering
  \includegraphics[width=1\linewidth]{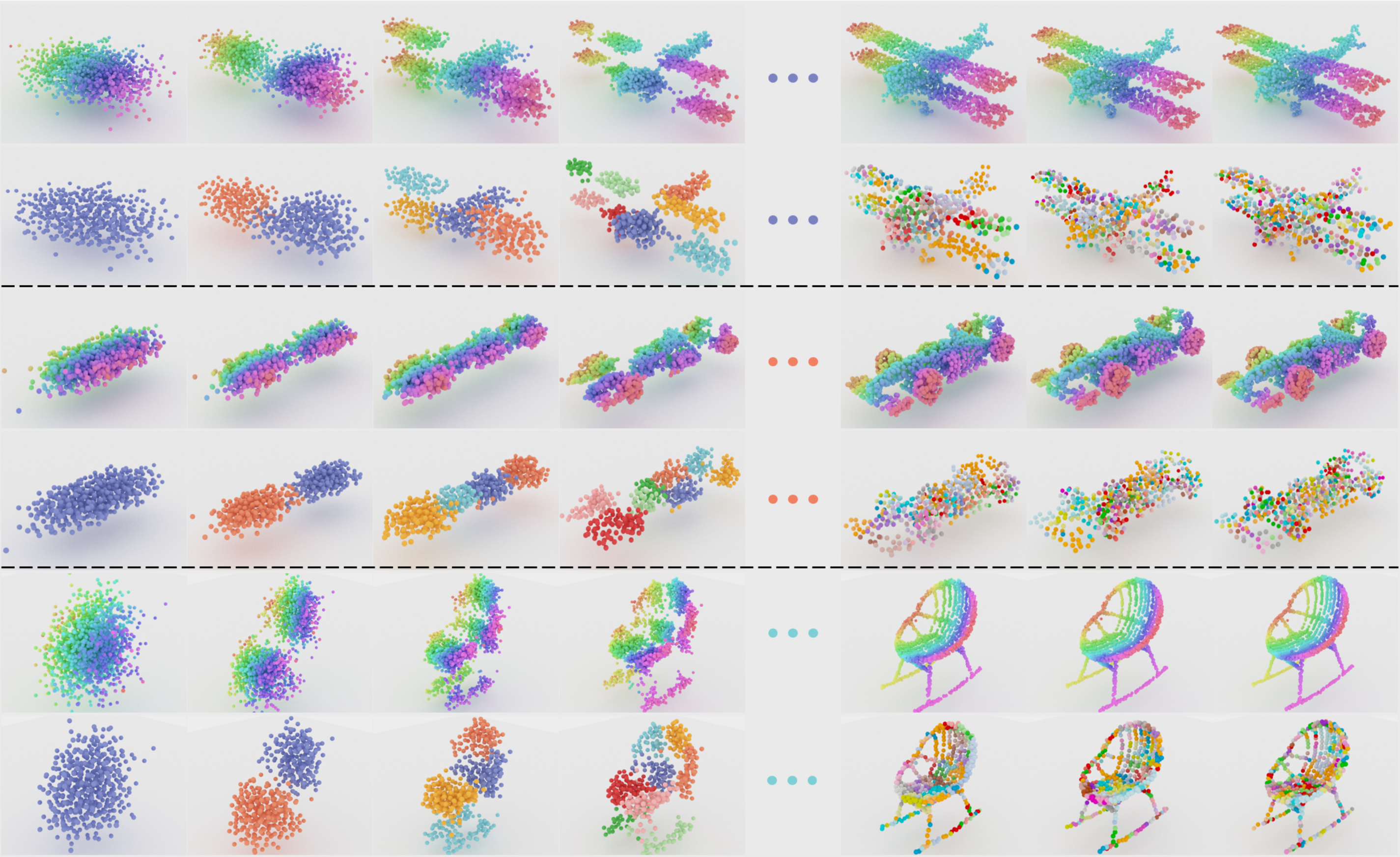}
  \caption{
  Qualitative visualization of the stage-wise cascaded generation process across the airplane, car, and chair categories. As the generation progresses from coarse to fine scales, the synthesized points (top row) are progressively anchored and refined by the evolving topology-aware geometric blueprint (bottom row), culminating in highly coherent and structurally precise 3D shapes.
  }
  \label{fig:4}
\end{figure}

\section{Experiments}

\subsection{Settings}

\noindent\textbf{Dataset.} 
Following established benchmarking protocols, we evaluate PointRSP on the ShapeNetV2 dataset, utilizing the preprocessing pipeline introduced by PointFlow~\cite{yang2019pointflow}. We focus on the three most structurally diverse and widely evaluated categories: airplanes, chairs, and cars. Each 3D shape is globally normalized within the range of $[-1.0, 1.0]$ and uniformly sampled to $2,048$ points. We adhere to the train/test splits in TIGER~\cite{ren2024tiger}, which comprise $2,832/405$, $4,612/662$, and $2,458/352$ for airplanes, chairs, and cars, respectively. All experiments strictly follow the same preprocessing strategy and data splits to ensure fair and consistent comparisons with the compared methods.

\noindent\textbf{Evaluation Metrics.} Consistent with the prior state-of-the-art models (e.g., PVD~\cite{zhou20213d} and LION~\cite{vahdat2022lion}), we adopt 1-nearest neighbor accuracy (1-NNA) ~\cite{lopez2016revisiting} as our primary quantitative metric. 
1-NNA rigorously assesses both the generation quality and the geometric diversity of the synthesized point clouds. A score approaching $50\%$ indicates that the generated distribution is virtually indistinguishable from the ground-truth reference distribution~\cite{yang2019pointflow}. To construct the 1-NN distance matrix, we compute two standard point cloud similarity measures: Chamfer Distance (CD) and Earth Mover's Distance (EMD). 
Furthermore, we employ Maximum Mean Discrepancy (MMD) to evaluate the global alignment between the generated and reference sets. In our ablation studies, we specifically report the standard deviation (Std.) of the MMD computed across batches. This batch-wise variance serves as a direct quantitative indicator of generation stability, allowing us to rigorously validate the efficacy of our proposed geometry-calibrated positional encoding in solving the "cold-start" issue.

\begin{table*}[t]
\centering
\caption{Quantitative evaluation of single-category generation on ShapeNetV2 for airplane, car, and chair, respectively.
The results are reported with 1-NNA (\%) under both CD and EMD metrics. The best and the second-best results are highlighted in \textbf{bold} and \underline{underline}, respectively.}
\label{tab:main_results}
\scriptsize
\begin{tabular}{
  >{\raggedright\arraybackslash}m{2.3cm} 
  >{\raggedright\arraybackslash}m{2.3cm} 
  >{\centering\arraybackslash}m{0.8cm} 
  >{\centering\arraybackslash}m{0.8cm} 
  >{\centering\arraybackslash}m{0.8cm} 
  >{\centering\arraybackslash}m{0.8cm}
  >{\centering\arraybackslash}m{0.8cm}
  >{\centering\arraybackslash}m{0.8cm}
  >{\centering\arraybackslash}m{0.8cm}
  >{\centering\arraybackslash}m{0.8cm}
}
\toprule
\multirow{2}{*}{\textbf{Model}} & \multirow{2}{*}{\textbf{Type}} 
& \multicolumn{2}{c}{\textbf{Airplane}} 
& \multicolumn{2}{c}{\textbf{Chair}} 
& \multicolumn{2}{c}{\textbf{Car}} & \textbf{Mean} & \textbf{Mean} \\
\cmidrule(lr){3-4} \cmidrule(lr){5-6} \cmidrule(lr){7-8}
 &  & \textbf{CD}  & \textbf{EMD}  
  & \textbf{CD}  & \textbf{EMD} 
  & \textbf{CD}  & \textbf{EMD} 
  & \textbf{CD}  & \textbf{EMD} \\
\midrule
1-GAN~\cite{achlioptas2018learning}            & GAN               & 87.30 & 93.95 & 68.58 & 83.84 & 66.49 & 88.78 & 74.12 & 88.86\\
PointFlow~\cite{yang2019pointflow}  & Normalizing Flow  & 75.68 & 70.74 & 62.84 & 60.57 & 58.10 & 56.25 & 65.54 & 62.52\\
DPM~\cite{luo2021diffusion}              & Diffusion         & 76.42 & 86.91 & 60.05 & 74.77 & 68.89 & 79.97 & 68.45 & 80.55\\
PVD~\cite{zhou20213d}              & Diffusion         & 73.82 & 64.81 & 56.26 & 53.32 & 54.55 & 53.83 & 61.54 & 57.32\\
LION~\cite{vahdat2022lion}            & Diffusion         & \underline{72.99} & 64.21 & 55.67 & 53.82 & 53.47 & \underline{53.21} & 61.75 & 57.59\\
TIGER~\cite{ren2024tiger}          & Diffusion         & 73.02 & \underline{64.10} & \underline{55.15} & \underline{53.18} & \underline{53.21} & 53.95 & 60.46 & 57.08\\
PointGrow~\cite{sun2020pointgrow}  & Autoregressive    & 82.20 & 78.54 & 63.14 & 61.87 & 67.56 & 65.89 & 70.96 & 68.77\\
CanonicalVAE~\cite{cheng2022autoregressive}
                            & Autoregressive    & 80.15 & 76.27 & 62.78 & 61.05 & 63.23 & 61.56 & 68.72 & 66.29\\
PointGPT~\cite{chen2023pointgpt}    & Autoregressive    & 74.85 & 65.61 & 57.24 & 55.01 & 55.91 & 54.24 & 63.44 & 62.24\\
PointNSP~\cite{meng2025pointnsp}           & Autoregressive    & 73.39 & 64.58 & 55.60 & 54.17 & 53.66 & 54.26 & 60.88 & 57.67\\
\textbf{PointRSP(Ours)}  & \textbf{Autoregressive} 
                            & \textbf{70.68} & \textbf{63.94} 
                            & \textbf{53.88} & \textbf{53.96} 
                            & \textbf{51.47} & \textbf{52.68} & \textbf{58.67} & \textbf{56.86}\\
\bottomrule
\end{tabular}
% \vspace{-5mm}
\end{table*}

\begin{figure}[t]
  \centering
  \includegraphics[width=1\linewidth]{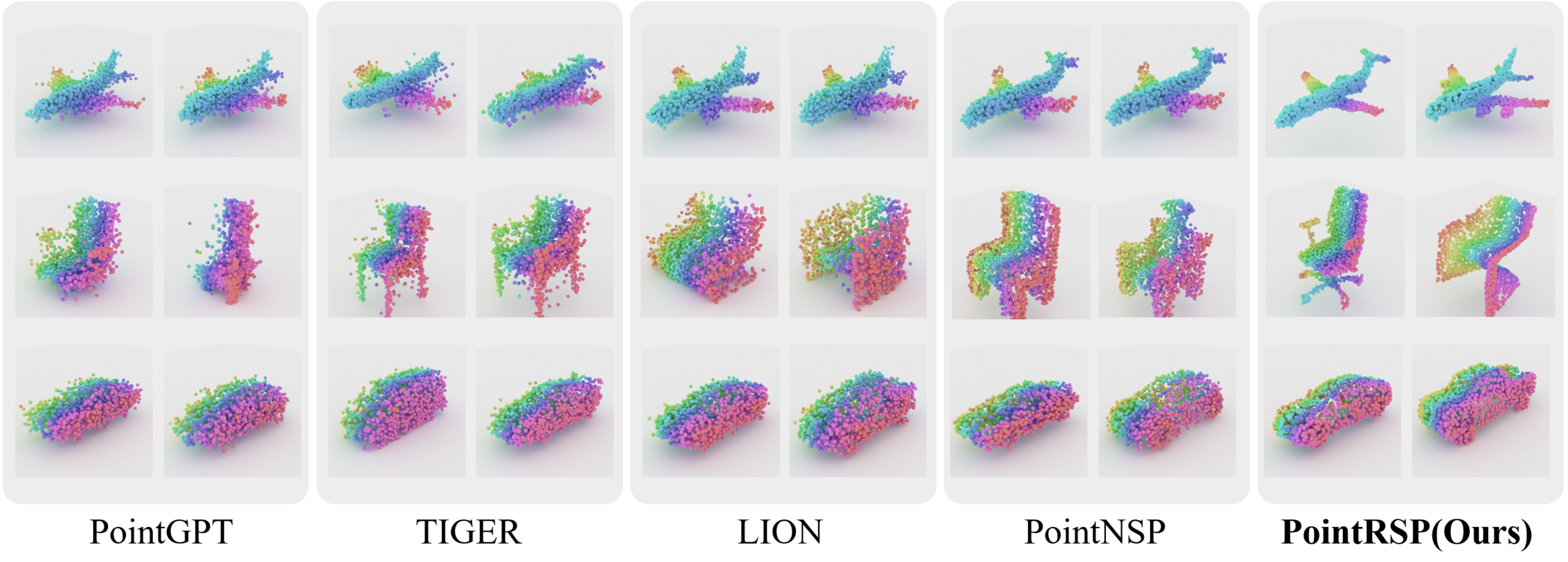}
  \caption{
  Qualitative comparison between PointRSP and representative state-of-the-art baselines across single-category generation. Compared to existing autoregressive and diffusion-based methods, PointRSP explicitly preserves intrinsic manifold continuity, resulting in synthesized 3D shapes with strictly maintained structural boundaries, fewer spatial outliers, and a highly coherent global geometry.
  }
  \label{fig:5}
\end{figure}

\subsection{Single-category Generation}\label{exp4_2}

\noindent\textbf{Quantitative Evaluation.} To rigorously evaluate PointRSP, we benchmark our framework against a comprehensive suite of recent state-of-the-art point cloud generation models. The baselines span both diffusion-based paradigms, including LION~\cite{vahdat2022lion} and TIGER~\cite{ren2024tiger}, and autoregressive approaches, such as PointGrow~\cite{sun2020pointgrow}, CanonicalVAE~\cite{cheng2022autoregressive}, PointGPT~\cite{chen2023pointgpt}, and PointNSP~\cite{meng2025pointnsp}. 
To ensure a fair comparison under the subset-scale paradigm, we instantiate the PointNSP baseline by adopting its official quantization-based VAE reconstruction module coupled with a lightweight global self-attention generator. 
As shown in Table~\ref{tab:main_results}, PointRSP consistently achieves state-of-the-art 1-NNA across all evaluated categories. This superiority over fragmented autoregressive models validates our motivation that heuristic spatial linearizations and stochastic downsampling inherently disrupt manifold continuity. Furthermore, our substantial improvements in the 1-NNA EMD metric, which is notoriously sensitive to local point density and fine-grained topological details, confirm the efficacy of our non-balanced binary tree representation. Unlike standard bisection strategies that force uniform spatial splits, our hybrid recursive spectral partitioning dynamically adapts to the intrinsic complexity of the 3D manifold, strictly preserving part boundaries. All experiments are conducted under a subset-scale configuration consistent with the shared transformer backbone setting used in our framework.

\noindent\textbf{Qualitative Evaluation.} 
Figure~\ref{fig:4} visualizes the cascaded generation process of PointRSP across different scales, explicitly rendering the intermediate point clouds alongside their corresponding structural labels. The visualization confirms that generated points consistently anchor around their predicted structural centers. As the geometric blueprint undergoes recursive spectral partitioning across progressive scales, the topological structure becomes systematically refined. This coarse-to-fine fidelity directly highlights the stability and geometric awareness inherent to our dual-stream cascaded design. Figure~\ref{fig:5} presents the visually comparison between PointRSP and representative competitors. Compared to existing methods, PointRSP synthesizes cleaner shapes characterized by fewer spatial outliers, strictly preserved structural boundaries, and a highly coherent global manifold. More visualizations are provided in the supplementary material.

\subsection{Multi-class Generation and Point Upsampling}

\noindent\textbf{Multi-Category Generation.}
Beyond the single-category setting, we investigate the scalability and expressiveness of PointRSP within a comprehensive multi-category generation paradigm. Following standard benchmarking protocols, we train the models with class conditioning across all $55$ ShapeNet object categories. This highly challenging setting requires the generator to simultaneously capture shared cross-category geometric regularities and preserve distinct, class-specific topological characteristics within a unified latent space. 

As detailed in the left forms of Table~\ref{tab:4_3}, PointRSP demonstrates exceptional generalization capability across these heterogeneous distributions. Specifically, under rigorous CD- and EMD-based evaluations, our framework achieves state-of-the-art performance in four of the experiments and preserves highly competitive results in the remaining two, demonstrating the superiority of PointRSP in models highly diverse 3D topologies without suffering from the manifold collapse. 

\noindent\textbf{Point Cloud Upsampling.}
Furthermore, we evaluate PointRSP on the downstream task of point cloud upsampling to validate its capacity for fine-grained geometric refinement. By conditioning the generator on a sparse input, we apply an upsampling factor of $2\times$, increasing the spatial resolution from $1,024$ to $2,048$ points. 
As shown in the right of Table~\ref{tab:4_3}, PointRSP explicitly outperforms all competing baseline approaches across every single evaluated category under both distance metrics. These definitive results indicate that our topology-preserving cascaded design not only excels in unconditional and class-conditional global generation but also seamlessly transfers to local geometric refinement. By anchoring generation to a deterministic structural blueprint, PointRSP accurately predicts highly localized structural details from coarse inputs. 

Extensive visualizations for both multi-category generation and upsampling are provided in the supplementary material.

\begin{table}[t]
\centering
\caption{Results of multi-class generation (left) and upsampling (right) on Airplane, Chair, and Car. Both are reported with 1-NNA (\%) under CD and EMD metrics. The best and the second-best results are highlighted in \textbf{bold} and \underline{underline}, respectively.}
\captionsetup[subtable]{labelformat=empty}
\scriptsize
\begin{subtable}[t]{0.48\linewidth}
  \centering
  \begin{tabular}{
      >{\raggedright\arraybackslash}m{1.2cm} 
      >{\centering\arraybackslash}m{0.7cm} 
      >{\centering\arraybackslash}m{0.7cm} 
      >{\centering\arraybackslash}m{0.7cm} 
      >{\centering\arraybackslash}m{0.7cm} 
      >{\centering\arraybackslash}m{0.7cm}
      >{\centering\arraybackslash}m{0.7cm}
  }
    \toprule
    \multirow{2}{*}{\textbf{Model}}
    & \multicolumn{2}{c}{\textbf{Airplane}} 
    & \multicolumn{2}{c}{\textbf{Chair}} 
    & \multicolumn{2}{c}{\textbf{Car}} \\
    \cmidrule(lr){2-3} \cmidrule(lr){4-5} \cmidrule(lr){6-7}
     & CD\tiny{$\downarrow$}  & EMD\tiny{$\downarrow$}
     & CD\tiny{$\downarrow$}  & EMD\tiny{$\downarrow$}
     & CD\tiny{$\downarrow$}  & EMD\tiny{$\downarrow$} \\
    \midrule
    LION & 86.30 & 77.04 & 66.50 & 63.83 & \underline{64.52} & 54.21 \\
    TIGER & 83.54 & 81.55 & \underline{57.34} & 61.45 & 65.79 & 57.24 \\
    PointGPT & 94.94 & 91.73 & 71.83 & 79.00 & 89.35 & 87.22 \\
    PointNSP & \underline{78.95} & \underline{68.84} & {58.79} & \textbf{55.10} & 69.97 & \textbf{52.89} \\
    \textbf{Ours} & \textbf{76.52} & \textbf{68.79} & \textbf{57.11} & \underline{56.07} & \textbf{58.39} & \underline{53.51} \\
    \bottomrule
  \end{tabular}
\end{subtable}\hfill
\begin{subtable}[t]{0.48\linewidth}
  \centering
    \begin{tabular}{
      >{\raggedright\arraybackslash}m{1.2cm} 
      >{\centering\arraybackslash}m{0.7cm} 
      >{\centering\arraybackslash}m{0.7cm} 
      >{\centering\arraybackslash}m{0.7cm} 
      >{\centering\arraybackslash}m{0.7cm} 
      >{\centering\arraybackslash}m{0.7cm}
      >{\centering\arraybackslash}m{0.7cm}
  }
    \toprule
    \multirow{2}{*}{\textbf{Model}}
    & \multicolumn{2}{c}{\textbf{Airplane}} 
    & \multicolumn{2}{c}{\textbf{Chair}} 
    & \multicolumn{2}{c}{\textbf{Car}} \\
    \cmidrule(lr){2-3} \cmidrule(lr){4-5} \cmidrule(lr){6-7}
     & CD\tiny{$\downarrow$}  & EMD\tiny{$\downarrow$}
     & CD\tiny{$\downarrow$}  & EMD\tiny{$\downarrow$}
     & CD\tiny{$\downarrow$}  & EMD\tiny{$\downarrow$} \\
    \midrule
    LION & 70.41 & \underline{59.65} & 53.98 & 54.33 & 57.14 & {47.56} \\
    TIGER & 71.65 & 59.94 & \underline{52.80} & \underline{52.98} & 57.90 & \underline{48.01} \\
    PointGPT & 72.11 & 60.12 & 53.75 & 53.21 & 57.26 & 47.85 \\
    PointNSP & \underline{69.86} & 59.68 & 53.79 & 53.84 & \underline{56.69} & 46.93 \\
    \textbf{Ours} & \textbf{67.29} & \textbf{59.09} & \textbf{52.05} & \textbf{52.63} & \textbf{54.95} & \textbf{48.63} \\
    \bottomrule
  \end{tabular}
\end{subtable}
\label{tab:4_3}
\end{table}

\begin{table}[t]
\centering
\caption{Ablation study on the single-category Car setting. MMD with its batch-wise Std. are reported to measure distribution alignment and stability. TPA and GCPE denote the Topology-Aware Partitioning Autoencoder and Geometry-Calibrated Positional Encoding, respectively. Cas./Struc. denote Cascaded and Structured processing.}
\scriptsize
\begin{tabular}{
  >{\centering\arraybackslash}m{0.8cm} 
  >{\centering\arraybackslash}m{0.8cm} 
  >{\centering\arraybackslash}m{0.8cm} 
  >{\centering\arraybackslash}m{0.9cm} 
  >{\centering\arraybackslash}m{0.9cm} 
  >{\centering\arraybackslash}m{1.5cm}
  >{\centering\arraybackslash}m{1.7cm}
  >{\centering\arraybackslash}m{1.7cm}
  >{\centering\arraybackslash}m{1.9cm}
}
\toprule
\multirow{2}{*}{\textbf{Model}} & \multicolumn{4}{c}{\textbf{Components}} & \multicolumn{4}{c}{\textbf{Results~$\downarrow$}} \\
\cmidrule(lr){2-5} \cmidrule(lr){6-9}
 & \textbf{Cas.} & \textbf{Struc.} & \textbf{TPA} & \textbf{GCPE} & 
$\mathrm{Acc.CD}_{\mathrm{NNA}}$ &
$\mathrm{Std.CD}_{\mathrm{MMD}}$ &
$\mathrm{Acc.EMD}_{\mathrm{NNA}}$ &
$\mathrm{Std.EMD}_{\mathrm{MMD}}$ \\
\midrule
Base &  &  &  &  & 53.66\tiny{\%} & 0.357\tiny{$\times10^{-3}$} & 54.26\tiny{\%} & 1.027\tiny{$\times10^{-3}$}\\
V1 & \ding{51} &  &  &  & 53.02\tiny{\%} & 0.343\tiny{$\times10^{-3}$} & 53.84\tiny{\%} & 0.923\tiny{$\times10^{-3}$} \\
V2 & \ding{51} & \ding{51} &  &  & 52.75\tiny{\%} & 0.318\tiny{$\times10^{-3}$} & 53.65\tiny{\%}  & 0.713\tiny{$\times10^{-3}$}\\
V3 & \ding{51} & \ding{51} & \ding{51} &  & 51.89\tiny{\%} & 0.304\tiny{$\times10^{-3}$} & 53.01\tiny{\%} & 0.636\tiny{$\times10^{-3}$} \\
Full & \ding{51} & \ding{51} & \ding{51} & \ding{51} & 51.47\tiny{\%} & 0.294\tiny{$\times10^{-3}$} & 52.68\tiny{\%} & 0.506\tiny{$\times10^{-3}$} \\
\bottomrule
\end{tabular}
\label{table:ab}
\end{table}

\subsection{Ablation Study}

To thoroughly quantify the effects of each proposed components, we conduct comprehensive ablation studies under the single-category Car generation setting, as detailed in Table~\ref{table:ab}. Following the description in Section~\ref{exp4_2}, we adopt the quantization-based VAE prototype of PointNSP~\cite{meng2025pointnsp} together with coupled with a lightweight global self-attention generator as our baseline.

Building upon the baseline, we first replace the standard fragmented generator with the cascaded generation framework (V1). Without the structural prior, the cascaded generation takes the low-resolution prediction as condition to generate high-resolution features. As evidenced by the reduced batch-wise standard deviation in MMD, the cascaded paradigm significantly stabilizes the generation process while preserving competitive fidelity. Next, we integrate a structural geometric prior. Emulating the NVG~\cite{wang2025next} approach (V2), we explicitly construct a full, balanced binary tree using the KNN clustering strategy. The results confirm that anchoring generation to a structural prior drastically mitigates the random fluctuations inherently caused by heuristic token orderings. However, forced uniform bisection remains suboptimal for 3D manifolds. By upgrading to our proposed Topology-Aware Partitioning Autoencoder driven by hybrid recursive spectral partitioning (V3), we explicitly accommodate the naturally non-balanced topological distribution of 3D point clouds. This topology-preserving blueprint prevents boundary blurring and leads to sharp improvements in 1-NNA. Finally, the full PointRSP framework integrates Geometry-Calibrated Positional Encoding, which explicitly resolves cold-start instability by anchoring early-stage latent features with multi-scale structural geometric centers, yielding our best generative performance.

Extensive hyperparameter analyses, including the effective spectral split horizon $T_k$ utilized in the TPA split stage and the positional encoding reconstruction weight $\alpha$, are provided in the supplementary material.

\section{Conclusion}

In this paper, we presented PointRSP, an autoregressive framework designed to reformulate 3D point cloud generation as a topology-preserving tessellation process. Our model employs a novel hybrid recursive spectral partitioning autoencoder to decompose point clouds into non-balanced binary trees, allowing the autoencoder to capture multiscale geometric dependencies within a quantized latent space. We further proposed a dual-stream generator to jointly model structural evolution and feature synthesis, supported by a geometry-calibrated positional encoding mechanism to stabilize early-stage generation. Extensive experiments demonstrate that PointRSP achieves state-of-the-art generation quality and diversity while exhibiting robust generalization on preserving topology across diverse 3D shapes.

\noindent\textbf{Limitations and Future Directions.} While PointRSP achieves promising results, it is currently limited to single-modality generation on object-centric shapes of standard densities, and the scalability of its recursive spectral partitioning mechanism to ultra-dense point sets or large-scale scene-level data has not been fully validated. Our future work will focus on incorporating multimodal conditional inputs such as text or images, and further optimizing the architecture to support the effective generation of large-scale, high-density point clouds.

\section{Acknowledgements}

This research is supported by National Natural Science Foundation of China (No.62302170), Guangdong Basic and Applied Basic Research Foundation (No.2024A1515010187), the Guangdong Natural Science Funds for Distinguished Young Scholars (Grant 2023B1515020097), the Singapore Ministry of Education Academic Research Fund Tier 2 (Award No. MOE-T2EP20125-0016), the Singapore Ministry of Education Academic Research Fund Tier 1 (Proposal ID: 24-SIS-SMU-015), and the Lee Kong Chian Fellowships.

\bibliographystyle{splncs04}
\bibliography{main}
\end{document}